\documentclass[11pt,a4paper]{article}
\usepackage[hyperref]{emnlp2020}
\usepackage{times}
\usepackage{latexsym}
\usepackage{graphicx}
\usepackage{subcaption}
\graphicspath{ {./figures/} }

\usepackage{amsmath}
\usepackage{tabularx}
\usepackage{xcolor}
\usepackage{booktabs}
\usepackage{mathtools}
\usepackage{bm}

\usepackage{microtype}

\aclfinalcopy 

\newcolumntype{Y}{>{\centering\arraybackslash}X}

\def\lc{\left\lfloor}   
\def\rc{\right\rfloor}

\title{The Secret is in the Spectra: Predicting Cross-lingual Task Performance with Spectral Similarity Measures}

\author{Haim Dubossarsky$^{\mathbf{1}}$ ~~ Ivan Vuli\'c$^{\mathbf{1}}$ ~~ Roi Reichart$^{\mathbf{2}}$ ~~ {Anna Korhonen}$^{\mathbf{1}}$\\
$^{\mathbf{1}}$ Language Technology Lab, University of Cambridge \\
$^{\mathbf{2}}$ Faculty of Industrial Engineering and Management, Technion, IIT\\
\texttt{\{hd423, iv250, alk23\}@cam.ac.uk} \hspace{0.8em} \texttt{roiri@ie.technion.ac.il} }

\date{}

\begin{document}
\maketitle
\begin{abstract}

Performance in cross-lingual NLP tasks is impacted by the (dis)similarity of languages at hand: e.g., previous work has suggested there is a connection between the expected success of bilingual lexicon induction (BLI) and the assumption of (approximate) isomorphism between monolingual embedding spaces. In this work we present a large-scale study focused on the correlations between monolingual embedding space similarity and task performance, covering thousands of language pairs and four different tasks: BLI,  parsing, POS tagging and MT. We hypothesize that statistics of the spectrum of each monolingual embedding space indicate how well they can be aligned. We then introduce several isomorphism measures between two embedding spaces, based on the relevant statistics of their individual spectra. We empirically show that \textbf{1)} language similarity scores derived from such spectral isomorphism measures are strongly associated with performance observed in different cross-lingual tasks, and \textbf{2)} our spectral-based measures consistently outperform previous standard isomorphism measures, while being computationally more tractable and easier to interpret. Finally, our measures capture complementary information to typologically driven language distance measures, and the combination of measures from the two families yields even higher task performance correlations. 
\end{abstract}

\section{Introduction}
\label{s:introduction}
The effectiveness of joint multilingual modeling and cross-lingual transfer in cross-lingual NLP is critically impacted by the actual languages in consideration \cite{Bender:2011lilt,Ponti:2019cl}. \textit{Characterizing, measuring, and understanding this cross-language variation} is often the first step towards the development of more robust multilingually applicable NLP technology \cite{Ohoran:2016coling,Bjerva:2019cl,Ponti:2019cl}. For instance, selecting suitable source languages is a prerequisite for successful cross-lingual transfer of dependency parsers or POS taggers \cite{Naseem:2012acl,Ponti:2018acl,deLhoneux:2018emnlp}. In another example, with all other factors kept similar (e.g., training data size, domain similarity), the quality of machine translation also depends heavily on the properties and language proximity of the actual language pair \cite{Kudugunta:2019emnlp}.

In this work, we contribute to this research endeavor by proposing a suite of spectral-based measures that capture the degree of isomorphism \cite{sogaard2018limitations} between the monolingual embedding spaces of  two languages. Our main hypothesis is that the potential to align two embedding spaces and learn transfer functions can be estimated through the differences between the monolingual embeddings' spectra. We therefore discuss representative statistics of the spectrum of an embedding space (i.e., the set of the singular values of the embedding matrix), such as its condition number or its sorted list of singular values. We then derive measures for the isomorphism between two embedding spaces based on these statistics.



To validate our hypothesis, we perform an extensive empirical evaluation with a range of cross-lingual NLP tasks. This analysis reveals that our proposed spectrum-based isomorphism measures better correlate and explain greater variance than previous isomorphism measures \cite{sogaard2018limitations,patra2019bilingual}. In addition, our measures also outperform standard approaches based on linguistic information \cite{littell-etal-2017-uriel}, 


The first part of our empirical analysis targets bilingual lexicon induction (BLI), a cross-lingual task that received plenty of attention, in particular as a case study to investigate the impact of cross-language variation on task performance \cite{sogaard2018limitations,Artetxe:2018acl}. Its popularity stems from its simple task formulation and reduced resource requirements, which makes it widely applicable across a large number of language pairs \cite{Ruder2018survey}. 

Prior work has empirically verified that for some language pairs BLI performs remarkably well, and for others rather poorly \cite{sogaard2018limitations,vulic2019we}. It attempted to explain this variance in performance by grounding it in the differences between the monolingual embedding spaces themselves. These studies introduced the notion of \textit{approximate isomorphism}, and argued that it is easier to learn a mapping function \cite{mikolov2013exploiting,Ruder2018survey} between language pairs whose embeddings are approximately isomorphic, than between languages pairs without this property \cite{miceli-barone-2016-towards,sogaard2018limitations}. Subsequently, novel methods to quantify the degree of isomorphism were proposed, and were shown to significantly correlate with BLI scores \cite{zhang2017earth, sogaard2018limitations, patra2019bilingual}. 

In this work, we report much higher correlations with BLI scores than existing isomorphism measures, across a variety of state-of-the-art BLI approaches. While previous work was limited only to coarse-grained analysis with a small number of language pairs (i.e., $<10$), our study is the first large-scale analysis that is focused on the relationship between quantifiable isomorphism and BLI performance. Our analysis covers hundreds of diverse language pairs, focusing on typologically, geographically and phylogenetically distant pairs as well as on similar languages. 



We further show that our findings generalize beyond BLI, to cross-lingual transfer in dependency parsing and POS tagging, and we also demonstrate strong correlations with machine translation (MT) performance. Finally, our spectral-based measures can be combined with typologically driven language distance measures to achieve further correlation improvements. This indicates the complementary nature of the implicit knowledge coded in continuous semantic spaces (and captured by our spectral measures) and the discrete linguistic information from typological databases (captured by the  typologically driven measures).

\section{Quantifying Isomorphism with Spectral Statistics}
\label{s:motivation}
Following the distributional hypothesis \cite{Harris:1954,Firth:1957}, word embedding models learn the meaning of words according to their co-occurrence patterns. Hence, the word embedding space of a language whose words are used in diverse contexts is  intuitively expected to encode richer information and greater variance than the word embedding space of a language with more restricting word usage patterns. 
The difference between two monolingual embedding spaces may also result from other reasons, such as the difference between the training corpora on which the embedding induction algorithm is trained, and the degree to which this algorithm accounts for the linguistic properties of each of the languages. 

While the exact combination of factors that govern the difference between the embedding spaces of different languages is hard to figure, this difference is likely to be indicative of the quality of cross-lingual transfer. This is particularly true when the embedding spaces are used by cross-lingual transfer algorithms. Our core hypothesis is that the difference between two monolingual spaces can be quantified by spectral statistics of the two spaces.

\subsection{Spectrum Statistics}
\label{ss:spectrum-stats}




Given a  $d$-dimensional embedding matrix $\mathbf{X}$, we perform Singular Value Decomposition (SVD) and obtain a diagonal matrix $\mathbf{\Sigma}$ whose main diagonal comprises of $d$ singular values, $\sigma_1, \sigma_2, \ldots, \sigma_d$, sorted in a descending order.\footnote{$\mathbf{X}$ is mean-centered, column means have been subtracted and are equal to zero.} Our aim is to quantify the difference between two embedding spaces by comparing statistics of their singular values. We next describe such statistics and in \S\ref{ss:measures} use them to measure the isomorphism between the spaces.

\vspace{1.8mm}
\noindent \textbf{Condition Number.} \label{ss:cn} 
In numerical analysis, a function's condition number measures the extent of change of the function's output value conditioned on a small change in the input \cite{blum14condition}. Consider the case of $\varphi:\mathbf{X}\rightarrow \mathbf{Y}$, where $\mathbf{X}$ and $\mathbf{Y}$ are two embedding spaces mapped via $\varphi$. The condition number, $\kappa(\mathbf{X})$, represents the degree to which small perturbations in the input $\mathbf{X}$ are amplified in the output $\varphi(X)$. Following \newcite{higham2015princeton}, we compute the condition number of an input matrix $\mathbf{X}$ with $d$ singular values as the ratio between its first (largest) and last (smallest) singular values:
%
{\normalsize
\begin{equation}
    \kappa(\mathbf{X})= {\frac{\sigma_1}{\sigma_d}}
    \label{Formula_condition}
\end{equation}}%
\noindent\textit{Why is it a relevant statistic?} A smaller condition number denotes a more ``stable'' matrix that is less sensitive to perturbations. Consequently, learning a transfer function $\varphi$ from one embedding space to another is more robust to noise when dealing with spaces with smaller $\kappa(\mathbf{X})$. We thus expect that embedding matrices with high condition numbers might impede the learning of good transfer functions in cross-lingual NLP: A function learnt on an embedding space that is sensitive to small perturbations may not generalize well.


\textit{Are small singular values reliable?} Small singular values are associated with noise, or with the least important information, and many noise reduction techniques remove them \cite{FORD2015299}. If the smallest singular value indeed captures noise, this might affect the condition number (Eq.~\eqref{Formula_condition}). It is thus crucial to distinguish between ``small but significant'' and ``small and insignificant'' singular values. This is what we do below.

\vspace{1.8mm}
\noindent \textbf{Effective Rank.}
\label{ss:erank} Given sorted singular values, how can we determine the last effective singular value? For a matrix with $d$ singular values $\sigma_1\geq\sigma_2\geq \cdots \geq\sigma_d\geq0$, the $\epsilon$-numerical rank can be defined as: $r_\epsilon= \min \{r:\sigma_r\geq\epsilon\}$, which means that singular values below a certain threshold are removed. However, this formulation introduces a dependency on the hyper-parameter $\epsilon$. To avoid this, \citet{roy2007effective} proposed an alternative method that considers the full spectrum of singular values before computing the so-called \textit{effective rank} of the input matrix \textbf{X}: 
\begin{align}
    erank(\textbf{X}) = \lc e^{H(\Sigma)} \rc
    \label{Formula_erank}
\end{align}
\noindent where $H(\Sigma)$ is the entropy of the matrix $\mathbf{X}$'s normalized singular value distribution $\bar{\sigma}_{i}=\frac{\sigma_i}{\sum_{i=1}^{d}\sigma_i}$ ,  computed as $H(\Sigma)=-\sum_{i=1}^{d} \bar{\sigma}_{i} \log \bar{\sigma}_{i}$. ${erank}(\mathbf{X})$, rounded to the smaller integer, yields the index of the last singular value that is considered significant, and is interpreted as the effective dimensionality, or rank, of the matrix $\mathbf{X}$. If $d$ is the dimensionality of the embedding space $\mathbf{X}$, and we assume that the number of word vectors in $\mathbf{X}$ is typically much larger than $d$, it then holds that (see \citet{roy2007effective}):
\begin{align}
    1 \leq erank(\mathbf{X}) \leq rank(\mathbf{X}) \leq d 
\end{align}
\noindent The dimensionality of an embedding space is intuitively assumed to be equal to the dimensionality of its constituent vectors: the matrix rank. Effective rank undermines this assumption: with effective rank matrices of the same `initial dimensionality' can have very different `true dimensionalities' \cite{yin2018dimensionality}. Effective rank is used for various problems outside NLP, such as source localization for acoustic \cite{tourbabin2015direction} and seismic  \cite{leeuwenburgh2014distance} waves, video compression \cite{bhaskaranand2010spectral}, and for the evaluation of implicit regularization in neural matrix factorization \cite{arora2019implicit}. We propose to use it to inform and improve the estimation of the condition number.





\vspace{1.8mm}
\noindent \textbf{Effective Condition Number.} We replace $\sigma_d$ in Eq.~\eqref{Formula_condition} with the singular value at the position of $\mathbf{X}$'s effective rank (see Eq.~\eqref{Formula_erank}), and compute the \textit{effective condition number} $\kappa_{ecn}$ as follows:
\begin{equation}
    \kappa_{ecn}(\mathbf{X})= {\frac{\sigma_1}{\sigma_{erank(\mathbf{X})}}}
    \label{Formula_effectivecn}
\end{equation}
\noindent In \S\ref{s:results} we empirically validate the  quality of the effective condition number in comparison to the standard condition number.

Having defined spectral statistics of an embedding space, we move to define means of comparing two spaces using these statistics.



\subsection{Spectral-Based Isomorphism Measures}
\label{ss:measures}
The statistics described in \S\ref{ss:spectrum-stats} capture properties of a single embedding space, but it is not straightforward how to employ them in order to quantify the similarity between two distinct embedding spaces. In what follows, we introduce isomorphism measures based on the spectral statistics.

Let us assume two embedding matrices $\mathbf{X}_1$ and $\mathbf{X}_2$, with their condition numbers, $\kappa(\mathbf{X_1})$ and $\kappa(\mathbf{X_2})$. We combine the two numbers using the harmonic mean function (\textsc{hm}) to derive an isomorphism measure between two embedding spaces, \textsc{cond-hm}($\mathbf{X}_1, \mathbf{X}_2$):
{\normalsize
\begin{align}
\textsc{cond-hm}(\mathbf{X}_1, \mathbf{X}_2) = \frac{2\cdot\kappa(\mathbf{X_1})\kappa(\mathbf{X_2})}{\kappa(\mathbf{X_1}) + \kappa(\mathbf{X_2})}
\end{align}}%
\noindent We similarly define the \textsc{econd-hm} measure over $\kappa_{ecn}(\mathbf{X_1})$ and $\kappa_{ecn}(\mathbf{X_2})$.

\noindent \textit{Why harmonic mean?} The higher the (effective) condition number of an embedding space, the higher its sensitivity to perturbations (i.e., the performance of transfer functions will be low). We view the condition number as a constraining factor on transferability, but what is the right way to evaluate the `transferability potential' of two spaces via their condition numbers? There are multiple ways to combine two condition numbers, but we have empirically validated (\S\ref{s:results}) that \textsc{hm} is a robust choice that outperforms some other possibilities (e.g., the arithmetic mean). We hypothesize this is because \textsc{hm} treats large discrepancies between two numbers in a manner that leans towards the smaller one (unlike e.g. arithmetic mean). Two noisy and two stable embedding spaces would have high and low \textsc{hm}s, respectively, but a noisy embedding space and a stable one would have an \textsc{hm} that leans towards the stable one.\footnote{Constraining factors are usually handled by taking the minimum (or maximum) values; however, this approach would impede the validity of our correlation analysis (see later \S\ref{s:experimental}) as it would artificially decrease the variance in only one variable (the language with the lowest measure will be systematically chosen). We refer the reader to Appendix C for an analysis.} Our results suggest that embedding spaces with small condition numbers can often tolerate noisy mappings from embedding spaces with high condition numbers, which might result from the improved stability of the former spaces.

\vspace{1.8mm}
\noindent \textbf{Singular Value Gap.}
In addition to \textsc{cond-hm} and \textsc{econd-hm}, we introduce another measure that empirically quantifies the divergence between the full spectral information of two embedding spaces. This measure quantifies the gap between the singular values obtained from the matrices $\textbf{X}_1$ and $\textbf{X}_2$ sorted in descending order. We define the measure of Singular Value Gap (\textsc{svg}) between two $d$-dimensional spaces $\mathbf{X}_1$ and $\mathbf{X}_2$, as the squared Euclidean distance between the corresponding sorted singular values after log transform:
\begin{align}
    SVG(\mathbf{X}_1, \mathbf{X}_2) = \sum_{i=1}^{d} ({\log\sigma^{1}_i} - {\log\sigma^{2}_i})^2
    \label{Formula_svg}
\end{align}
\noindent where $\sigma^{1}_i$ and $\sigma^{2}_i$, $i=1,\ldots,d$ are the sorted singular values characterizing the two embedding matrices $\mathbf{X}_1$ and $\mathbf{X}_2$. The intuition here is that two embedding spaces with similar singular values at the same index will be more isomorphic and therefore easier to align into a shared space, and enable more effective cross-lingual transfer. 





In summary, this section has presented methods that estimate the degree of isomorphism between any given pair of embedding spaces, which may differ in their language, training corpus, embedding induction algorithm or in other factors. While the focus of our empirical analysis (\S\ref{s:experimental}, \S\ref{s:results}) is cross-language learning and transfer, we note that the scope of our methods may be wider, and that they have not been developed only with cross-lingual learning in mind.

\section{Related Work and Baselines}
\label{s:related}
We now provide an overview of prior research that focused on two relevant themes: \textbf{1)} measuring approximate isomorphism between two embedding spaces, and \textbf{2)} more generally, quantifying the (dis)similarity between languages, going beyond isomorphism measures. The discussed approaches will also be used as the main baselines later in \S\ref{s:results}.

\vspace{1.8mm}
\noindent \textbf{Measuring Approximate Isomorphism.} 
We focus on two standard isomorphism measures from prior work which are most similar to our work, and use them as our main baselines. The first measure, termed \textbf{Isospectrality (IS)} \cite{sogaard2018limitations}, is based on spectral analysis as well, but of the Laplacian eigenvalues of the nearest neighborhood graphs that originate from the initial embedding spaces $\mathbf{X}_1$ and $\mathbf{X}_2$ (for further technical details see Appendix A). \newcite{sogaard2018limitations} argue that these eigenvalues are compact representations of the graph Laplacian, and that their comparison reveals the degree of (approximate) isomorphism. Although similar in spirit to our approach, constructing nearest neighborhood graphs (and then analyzing their eigenvalues) removes useful information on the interaction between all vectors from the initial space, which our spectral method retains.

The second measure is the \textbf{Gromov-Hausdorff distance (GH)} introduced by \newcite{patra2019bilingual}. It measures the maximum distance of a set of points to the nearest point in another set, or in other words the worst case distance between two metric spaces $\mathcal{X}$ and $\mathcal{Y}$ (for further technical details see again Appendix A). \citet{patra2019bilingual} propose this distance to test how well two language embedding spaces can be aligned under an isometric transformation.

While both IS and GH were reported to have strong correlations with BLI performance in prior work, they have not been evaluated in large-scale experiments before. In fact, the correlations were computed on a very small number of language pairs (IS: 8 pairs, GH: 10 pairs). Further, both measures do not scale well computationally. Therefore, for computational tractability, the scores are computed only on the sub-matrices spanning the sub-spaces of the most frequent subsets from the full embedding spaces (IS: 10k words, GH: 5k words). In this work, we provide full-fledged empirical analyses of the two measures on a much larger number of pairs from diverse languages, and compare them against the spectral-based measures introduced in \S\ref{s:motivation}. The fact that the proposed spectral-based methods are grounded in linear algebra theory (cf. \S\ref{ss:spectrum-stats}) also arguably provides a more intuitive understanding of their theoretical underpinning than what is currently offered in the relevant prior work.






\vspace{1.8mm}
\noindent \textbf{Measuring Language Similarity.}
At the same time, distances between language pairs can also be captured through (dis)similarities in their \textit{discrete} linguistic properties, such as overlap in syntactic features, or proximity along the phylogenetic language tree. The properties are typically hand-crafted, and are extracted from available typological databases such as the World Atlas of Languages (WALS) \cite{wals} or URIEL \cite{littell-etal-2017-uriel}, among others \cite{Ohoran:2016coling,Ponti:2019cl}. Such distances were found useful in guiding and informing cross-lingual transfer tasks \cite{cotterell-heigold-2017-cross,Agic:2017udw,lin-etal-2019-choosing,Ponti:2019cl}. 

In particular, we compare against three precomputed measures of language distance based on the URIEL typological database \cite{littell-etal-2017-uriel}. \textbf{Phylogenetic distance (PHY)} is derived from the hypothesized phylogenetic tree of language descent. \textbf{Typological distance (TYP)} is computed based on the overlap in syntactic features of languages from the WALS database \cite{wals}. \textbf{Geographic distance (GEO)} is obtained from the locations where languages are spoken; see the work of \newcite{littell-etal-2017-uriel} for more details.

We use these isomorphism measures and linguistic measures as language distance measures. We simply compute language distance between two languages $L_1$ and $L_2$ as $LDist(L_1, L_2)=D(\mathbf{X}_1, \mathbf{X}_2)$, where $D=\{$\textsc{svg}, \textsc{cond-hm}, \textsc{econd-hm}, \textsc{gh}, \textsc{is}, \textsc{phy}, \textsc{typ}, \textsc{geo}$\}$. Later in \S\ref{s:results} we show that ``proxy'' language distances originating from the proposed spectral-based isomorphism measures (see \S\ref{ss:measures}) correlate better with cross-lingual transfer scores across several tasks, than these language distances which are based on discrete linguistic properties. We verify that implicit knowledge coded in continuous embedding spaces and linguistic knowledge explicitly coded in external databases often complement each other.



\section{Experimental Setup}
\label{s:experimental}
The conducted empirical analyses can be divided into two major parts. First, we run large-scale BLI analyses across several hundred language pairs from dozens of languages, comparing the correlation of spectral-based isomorphism measures (\S\ref{ss:measures}) and all baselines (\S\ref{s:related}) with performance of a wide spectrum of state-of-the-art BLI methods. Second, we run further correlation analyses with performances in cross-lingual downstream tasks: dependency parsing, POS tagging, and MT. We first provide the details of the experiments that are shared between the two parts, and then provide further specifics of each experimental part.


\vspace{1.8mm}
\noindent \textbf{Monolingual Word Embeddings.} 
For all isomorphism measures (\textsc{svg}, \textsc{cond-hm}, \textsc{econd-hm}, \textsc{gh} and \textsc{is}) and languages in our analyses we use publicly available 300-dim monolingual fastText word embeddings pretrained on Wikipedia with exactly the same default settings (see \newcite{bojanowski2017enriching}), length-normalized and trimmed to the 200k most frequent words.\footnote{\textcolor{darkblue}{fasttext.cc/docs/en/pretrained-vectors.html}}


\vspace{1.8mm}
\noindent \textbf{Isomorphism Measures: Technical Details.} For our spectral-based measures, we compute a full SVD decomposition (i.e., no dimensionality reduction) of the embedding space. We compute \textsc{svg} scores for BLI based on the first 40 singular values, which we empirically found to produce slightly better results;\footnote{The average gains were around 10\%; we note that running \textsc{svg} with the full set of singular values also outperforms the baseline measures.} for the other tasks we use all singular values. For \textsc{is} and \textsc{gh}, we replicate the experimental setup from prior work: we compute the \textsc{is} score over the top 10k most frequent words in each monolingual space, while the \textsc{gh} score is computed over the top 5k words from each monolingual space.\footnote{\url{github.com/joelmoniz/BLISS}}

\subsection{Bilingual Lexicon Induction}
\label{ss:bli}
We conduct correlation analyses of the results from previous studies that report BLI scores for a large number of language pairs. On top of that, we complement the existing results from previous research with new results obtained with state-of-the-art BLI methods, applied to additional language pairs.

\vspace{1.8mm}
\noindent \textbf{BLI Setups and Scores.}
\citet{vulic2019we} ran BLI experiments on 210 language pairs, spanning 15 diverse languages. Their training and test dictionaries (5k and 2k translation pairs) are derived from \textbf{PanLex} \cite{Baldwin2010panlex,Kamholz2014panlex}. We complement the original 210 pairs with additional 210 language pairs of 15 closely related (European) languages using dictionaries extracted from PanLex following the procedure of \newcite{vulic2019we}. With the additional language set, the aim is to probe if isomorphism measures can also capture more subtle and smaller language differences.\footnote{The initial set of \newcite{vulic2019we} comprises Bulgarian, Catalan, Esperanto, Estonian, Basque, Finnish, Hebrew, Hungarian, Indonesian, Georgian, Korean, Lithuanian, Norwegian, Thai, Turkish. The additional 210 language pairs are only composed of Germanic, Romance and Slavic languages. For a full list of the languages see Table 4 in the appendix.} 



We also analyze the BLI results of 108 language pairs from \textbf{MUSE} \cite{conneau2017word}. This dataset systematically covers English, with 88 translation pairs that involve English as either the source or target language. Finally, we analyze the available BLI results of  \citet{glavas-etal-2019-properly} (referred to as \textbf{GTrans}) that are based on dictionaries obtained from Google Translate and include 28 language pairs spanning 8 different languages. For the full list of language pairs involved in previous BLI studies, we refer the reader to prior work \cite{conneau2017word,glavas-etal-2019-properly,vulic2019we}.

\vspace{1.8mm}
\noindent \textbf{BLI Methods in Comparison.} The scores in each BLI setup were computed by several state-of-the-art BLI methods based on cross-lingual word embeddings, briefly described here. \textbf{1)} \textsc{Sup} is the standard supervised method \cite{artetxe2016learning,smith2017offline} that learns a mapping between two embedding spaces based on a training dictionary by solving the orthogonal Procrustes problem \cite{schonemann1966generalized}. \textbf{2)} \textsc{Sup+} is another standard supervised method that additionally applies a variety of pre-processing and post-processing steps (e.g., whitening, dewhitening, symmetric re-weighting) before and after learning the mapping matrix, see \cite{Artetxe:2018acl}. \textbf{3)} \textsc{Unsup} is a fully unsupervised method based on the ``similarity of monolingual similarities'' heuristic to extract the seed dictionary from monolingual data. It then uses an iterative self-learning procedure to improve on the initial noisy dictionary \cite{Artetxe:2018acl}. For more technical details on the fully unsupervised model, we refer the reader to prior work \cite{Ruder:2019tutorial,vulic2019we}.\footnote{The \textsc{Sup+} and \textsc{Unsup} methods are based on the VecMap framework (\url{github.com/artetxem/vecmap}) which showed very competitive and robust BLI performance across a wide range of language pairs in recent comparative analyses \cite{glavas-etal-2019-properly,vulic2019we,doval2019onthe}.}

In sum, our analyses are conducted in three BLI setups (PanLex, MUSE, GTrans) and examine three types of state-of-the-art mapping-based methods, both supervised and unsupervised (\textsc{Sup}, \textsc{Sup+}, \textsc{Unsup}). Altogether, these span 556 language pairs, and cover both related and distant languages.\footnote{We report all results for each BLI method, dictionary and language pairs in the supplementary material (and also here \url{https://tinyurl.com/skn5cf7}). We also report scores with another method, RCSLS \cite{Joulin:2018emnlp}, benchmarked in the GTrans BLI setup (see Table~\ref{tab:full_BLI_table}).} Following prior work \cite{glavas-etal-2019-properly}, our BLI evaluation measure is Mean Reciprocal Rank (MRR). We note that identical findings emerge from running the correlation analyses based on Precision@1 scores in lieu of MRR.




\subsection{Downstream Tasks}
\label{ss:downstream}

Following the large-scale nature of our BLI analyses, we run similar correlation analyses on several downstream tasks that comprise a large number of (both similar and distant) language pairs.\footnote{For the full list of languages that were analyzed throughout all our experiments see Table \ref{tab:language_counts} in the appendix.} We rely on results from a recent study of \citet{lin-etal-2019-choosing} that focused on cross-lingual transfer performance in MT, dependency parsing, and POS tagging.\footnote{For full details regarding the models used to compute the scores for each downstream task, we refer the interested reader to the work of \newcite{lin-etal-2019-choosing} and the accompanying repository: \url{https://github.com/neulab/langrank}. We note that scores for each language pair in each task have been produced with the same task architectures.}




\vspace{1.8mm}
\noindent \textbf{Machine Translation.}
\citet{lin-etal-2019-choosing} report BLEU scores when translating 54 source $L_1$ languages into English as the target language. We report correlations between the different language distance measures and these 54 BLEU scores.



\vspace{1.8mm}
\noindent \textbf{Dependency Parsing.}
We base our analysis on the cross-lingual zero-shot parser transfer results of \newcite{lin-etal-2019-choosing}: The standard biaffine dependency parser \cite{dozat2017deep,dozat-etal-2017-stanfords} is trained on the training portions of Universal Dependencies (UD) treebanks from 31 languages \cite{nivre2018ud23}, and is then used to parse the test treebank of each language, now used as the target language. We report correlations between the language distance measures and the Labeled Attachment Scores (LAS) for all combinations of 31 languages, resulting in 930 pairs.


\vspace{1.8mm}
\noindent \textbf{POS Tagging.} We use POS tagging accuracy scores reported by \citet{lin-etal-2019-choosing}. These scores span 26 low-resource target languages and 60 source languages which measure the utility of each source language to each of the 26 target languages in POS tagging. We use a sample of 840 language pairs for the correlation analysis, as 16 low-resource target languages and 49 source languages have readily available pretrained fastText vectors.




\subsection{Correlation Analyses and Statistical Tests}
\label{ss:statanalysis}


All scores from isomorphism measures and BLI scores were log-transformed\footnote{This is an order-preserving transformation which is often used when the data distribution is skewed (as was the case with our BLI and isomorphism scores).} prior to any correlation computation. We report Pearson's correlation coefficients in all tasks. This allows us to investigate which of the different individual measures is most important to predict task performance.


\vspace{1.8mm}
\noindent \textbf{Regression Analyses.} The individual (i.e., single-variable) analyses are not sufficient to account for the complex interdependencies between the distance measures themselves, and how they interact with task performance when combined. Therefore, we also use standard linear stepwise regression model \cite{Hocking:1976,Draper:1998regression}:
{\normalsize
\begin{align}
\mathbf{Y} = \beta_0 + \beta_1 x_1 + \ldots + \beta_n x_n + \epsilon . 
\end{align}}%
\noindent Here, task performance $\mathbf{Y}$ is predicted using a set of regressors, $x_1,\ldots,x_n$ (i.e., \textsc{svg}, \textsc{cond-hm}, \textsc{econd-hm}, \textsc{gh}, \textsc{is}, \textsc{phy}, \textsc{typ}, \textsc{geo}), that are added to the model incrementally only if their marginal addition to predicting $\mathbf{Y}$ is statistically significant ({$p<.01$}). This method is useful for finding variables (i.e., in our case distance measures) with \textit{maximal} and \textit{unique} contribution to the explanation of $\mathbf{Y}$, when the variables themselves are strongly cross-correlated, as in our case. 

This model is able to: (a) discern which variables overlap in their information; (b) detect variables that complement each other; and (c) evaluate their joint contribution in predicting task performance. We compute the regression model's score for all statistically significant variables, and report its square-root, $\hat{r}$. 

Importantly, $\hat{r}$ is \textit{not} a one-number description of a language, but rather an illustrative quantification of the joint contribution of several \textit{different} distance measures to the explanation of $\mathbf{Y}$. Its goal is to investigate potential gains achieved through the combination of several distance measures, as opposed to using a single-best measure. The distance measures that are found statistically significant in the regression analyses are marked by superscripts over $\hat{r}$ (see later in Tables~\ref{tab:full_BLI_table} and \ref{tab:downstream_table}).



\section{Analyses and Results}
\label{s:results}
\begin{table*}[t!]
\centering
\def\arraystretch{0.999}
{\footnotesize
\begin{tabularx}{\linewidth}{ll YYYYYYYY}
\toprule 
{} & {} & \textbf{PanLex} & {} & {} & \textbf{MUSE} & {} & {} & \textbf{GTrans} & {} \\ 
{} & \textsc{Sup*} & \textsc{Sup+} & \textsc{Unsup} & \textsc{Sup} & \textsc{Sup+} & \textsc{Unsup} & \textsc{Sup} & \textsc{rcsls} & \textsc{Unsup} \\ 
\midrule
{1. \textsc{svg}} & {-.74} & {-.76} & {\bf -.74} & {\textbf{-.60}} & {-.60} & {\textbf{-.61}} & {-.59} & {-.59} & {-.53} \\
{2. \textsc{cond-hm}} & {-.66} & {-.64} & {-.56} & {-.58} & {-.47} & {-.39} & {-.46} & {-.47} & {-.42} \\
{3. \textsc{econd-hm}} & {\bf -.79} & {\bf -.77} & {-.70} & {-.46} & {-.38} & {-.39} & {\bf -.74} & {\bf -.75} & {\bf -.56} \\
\cmidrule(lr){1-10}
{4. \textsc{gh}} & {-.41} & {-.41} & {-.46} & {-.43} & {-.33} & {-.28} & {-.47} & {-.46} & {-.50} \\
{5. \textsc{is}} & {-.51} & {-.53} & {-.49} & {-.42} & {-.42} & {-.34}  & {-.33} & {-.33} & {-.36} \\
\cmidrule(lr){1-10}
{6. \textsc{phy}} & {-.55} & {-.55} & {-.41} & {-.53} & {-.62} & {-.34} & {-.51} & {-.50} & {-.52} \\
{7. \textsc{typ}} & {-.57} & {-.52} & {-.38} & {-.47} & {\bf -.64} & {-.44} & {-.59} & {-.59} & {-.43} \\
{8. \textsc{geo}} & {-.62} & {-.66} & {-.62} & {-.57} & {-.54} & {-.35} & {-.22} & {-.22} & {-.35} \\
\cmidrule(lr){1-10}
{$\hat{r}$} & {.91$^{1,3,6-8}$} & {.91$^{1,3,6-8}$} & {.82$^{1,3,8}$} & {.69$^{1,6}$} & {.74$^{7,8}$} & {.61$^1$} & {.92$^{3,6,8}$} & {.93$^{3,6,8}$} & {.86$^{3,6,8}$} \\
\bottomrule
\end{tabularx}}%
\vspace{-0mm}
\caption{Correlations with BLI performance in three BLI setups, see \S\ref{ss:bli}. The best distance measure for each setup and BLI method is \textbf{bolded}.  $\hat{r}$ is the score from the stepwise regression model, see \S\ref{ss:statanalysis}. Superscripts indicate the distance measures that are statistically significant and included in the stepwise regression model (e.g., .91$^{1,3,6-8}$ means: \textsc{svg}, \textsc{econd-hm} and all the linguistic distances have a combined contribution equivalent to 0.91 Pearson). *See the scatter plot in Appendix C.}

\label{tab:full_BLI_table}
\end{table*}

\begin{table}[!t]
\centering
\def\arraystretch{0.999}
{\footnotesize
\begin{tabularx}{\linewidth}{l YYY}
\toprule 
{} & \textbf{MT} & \textbf{DEP} & \textbf{POS}  \\ 
\midrule
{1. \textsc{svg}} & {-.56} & {\textbf{-.79}} & {\textbf{-.52}} \\
{2. \textsc{cond-hm}} & {-.52} & {-.60} & {-.41} \\
{3. \textsc{econd-hm}} & {-.51} & {-.66} & {-.44} \\
\cmidrule(lr){1-4}
{4. \textsc{gh}} & {-.16}  & {-.70} & {\textbf{-.52}} \\
{5. \textsc{is}} & {-.13}  & {-.67} & {-.43} \\
\cmidrule(lr){1-4}
{6. \textsc{phy}} & {-.45}  & {-.56} & {-.16} \\
{7. \textsc{typ}} & {\textbf{-.66}} & {-.40} & {-.16} \\
{8. \textsc{geo}} & {-.26} & {-.58} & {-.15} \\
\cmidrule(lr){1-4}
{$\hat{r}$} & {.74$^{1,7}$} & {.87$^{1,6-8}$} & {.59$^{1,4,5,7}$}  \\
\bottomrule
\end{tabularx}}%
\vspace{-0mm}
\caption{Correlations with performance in three other cross-lingual tasks: Machine Translation (MT), dependency parsing (DEP), and POS tagging. Results for the best distance measure are highlighted in \textbf{bold}. $\hat{r}$ is computed using the stepwise regression model (see \S\ref{ss:statanalysis}). 
}
\label{tab:downstream_table}
\end{table}

The results are summarized in Tables~\ref{tab:full_BLI_table} and \ref{tab:downstream_table}. The first main finding is that our proposed spectral-based isomorphism measures are strongly correlated with performance across all tasks and settings.\footnote{The negative correlations between \textsc{svg} and \textsc{econd-hm} scores and task performance have a clean interpretation: they are distance/dissimilarity measures, so high scores of those measures indicate less similar languages.} In fact, they show the strongest individual correlations with task performance among all isomorphism measures and linguistic distances alike. The only exception is the MT task, where our measures fall short of \textsc{typ} (see Table~\ref{tab:downstream_table}), although we mark that they still hold a strong advantage over the baseline \textsc{gh} and \textsc{is} isomorphism measures that do not seem to capture any useful language similarity properties needed for the MT task. 

\textsc{econd-hm} systematically outperforms \textsc{cond-hm} on 2 of 3 BLI datasets and 2 of 3 downstream tasks, validating our assumption that discarding the smallest singular values reduces noise. Additionally, \textsc{svg} shows greater stability across tasks and datasets than \textsc{econd-hm}. A general finding across all tasks is that our spectral measures are the most robust isomorphism measures: they substantially outperform the widely used baselines \textsc{gh} and \textsc{is}.






As stepwise regression discerns between overlapping and complementing variables (see \S\ref{ss:statanalysis}), another finding indicates that our spectral measures are complemented by linguistically driven language distances. Indeed, their combination achieves very high correlation scores. The results demonstrate this across all tasks and settings (see bottom rows of the tables). For instance, when combining spectral measures with the linguistic distances, the regression model reaches outstanding correlation scores up to $r=.91$ on PanLex BLI (Table~\ref{tab:full_BLI_table}); with 420 language pairs, PanLex is the most comprehensive BLI dataset in our study. In addition, \textsc{gh} and \textsc{is} are not chosen as significant regressors in the stepwise regression model, which indicates that they capture less information than our spectral methods.\footnote{This does not imply \textsc{gh} or \textsc{is} do not capture information that is complementary to discrete linguistic information, only that the implicit knowledge captured by \textsc{gh} or \textsc{is} cannot match that of our spectral-based methods, and their combination with linguistic distances obtained from external knowledge sources.} Overall, the regression results support the notion that conceptually different distances (i.e., isomorphism-based versus measures based on linguistic properties) capture different properties of similarities between languages, which has a synergistic effect when they are combined.


Concerning individual tasks, we note that our spectral-based measures outperform the baselines regardless of the underlying BLI method. Further, \textsc{svg} is the most informative distance measure in parsing experiments, and all linguistic distances fall behind isomorphism measures. Combining linguistic distances with \textsc{svg} increases the already high correlation from 0.79 to 0.87 (Table \ref{tab:downstream_table}, second column). For the POS tagging task, \textsc{svg} and \textsc{gh} are on par, and the combination with \textsc{is} and \textsc{typ} increase their correlation from 0.52 to 0.59 (Table \ref{tab:downstream_table}, third column). This is the only analysis where baseline methods are found significant.

Additional results and analyses are provided in Appendix B. They further demonstrate that our measures also indicate transfer quality of different target languages for a given source language, and transfer quality of source languages for a given target language, for the tasks discussed in this paper.

\section{Further Discussion and Conclusion}
\label{s:discussion}



This work introduces two spectral-based measures, \textsc{svg} and \textsc{econd-hm}, that excel in predicting performance on a variety of cross-lingual tasks. Both measures leverage information from singular values in different ways: \textsc{econd-hm} uses the ratio between two singular values, and is grounded in linear algebra and numerical analysis \cite{blum14condition,roy2007effective}, while \textsc{svg} directly utilizes the full range of singular values. We suspect that the use of the full range of singular values is what makes \textsc{svg} more robust across different tasks and datasets, compared to \textsc{econd-hm} that shows greater variance, as observed in our results above. 

While the spectral methods are computed solely on word vectors from Wikipedia, the results in the downstream tasks are computed with different sets of embeddings (e.g., multilingual embeddings for dependency parsing), or the embeddings are learnt during training (for POS tagging and MT). We believe that this discrepancy does not constitute a shortcoming of our analyses, but rather the opposite: spectral-based methods maintain their high correlations in the downstream tasks as well, and this supports the notion that these measures might indeed capture deeper linguistic information than mere similarities between embedding spaces.


%

Our use of effective rank in improving the condition number (via effective condition number) is also inspired by recent work that aimed to automatically detect true dimensionality of embedding spaces. However, previous work has taken an empirical approach by simply tuning embedding dimensionality to evaluation tasks at hand \cite{wang2019single,raunak2019effective,carrington2019invariance}. Our intention, on the other hand, is to extract the true embedding dimensionality directly from the embedding space. Another recent study \cite{yin2018dimensionality} employed perturbation analysis to study the robustness of embedding spaces to noise in monolingual settings, and established that it is also related to effective dimensionality of the embedding space. All these inspired us to replace the standard matrix rank with effective rank when computing the condition number, and to introduce the statistic of effective condition number in \S\ref{ss:spectrum-stats}.

Our study is also the first to compare language distance measures that are based on discrete linguistic information \cite{littell-etal-2017-uriel} with measures of isomorphism (i.e., our spectral-based measures, \textsc{is}, \textsc{gh}), which can also be used as proxy language distance measures. Our findings, suggesting that it is possible to effectively combine these two types of language distance measures, call for further research that will advance our understanding of: 1) what knowledge is captured in monolingual and cross-lingual embedding spaces \cite{Gerz:2018tacl,Pires:2019acl,Artetxe2019cross-lingual}; 2) how that knowledge complements or overlaps with linguistic knowledge compiled into lexical-semantic and typological databases \cite{wals,Wichmann:2018asjp,Ponti:2019cl}; and 3) how to use the combined knowledge for more effective transfer in cross-lingual NLP applications \cite{Ponti:2018acl,Eisenschlos2019multifit}.

The differences in embedding spaces of different languages do not only depend on linguistic properties of the languages in consideration, but also on other factors such as the chosen training algorithm, underlying training domain, or training data size and quality \cite{sogaard2018limitations,arora2019implicit,vulic2020all}. In future research we also plan an in-depth study of these factors and their relation to our spectral analysis.


We believe that the main insights from this study will inform and guide different cross-lingual transfer learning methods and scenarios in future work. These might range from choosing source languages for transfer in low-data regimes, over monolingual word vector induction guided by the spectral statistics, even to more effective hyperparameter search.

\section*{Acknowledgments}
The work of IV and AK is supported by the ERC Consolidator Grant LEXICAL: Lexical Acquisition Across Languages (no 648909) awarded to AK. HD is supported by the Blavatnik Postdoctoral Fellowship Programme. HD would like to thank Dr. Avital Hahamy for her contribution to the initiation of this work.

\bibliography{emnlp2020}
\bibliographystyle{acl_natbib}

\clearpage
\appendix
\section{IS and GH}
\label{s:appendix_a}

\noindent\textbf{Isospectrality (IS)} After length-normalizing the vectors, \newcite{sogaard2018limitations} compute the nearest neighbor graphs using a subset of the top $N$ most frequent words in each space, and then calculate the Laplacian matrices $LP_1$ and $LP_2$ of each graph. For $LP_1$, the smallest $k_1$ is then sought such that the sum of its $k_1$ largest eigenvalues $\sum^{k_1}_{i=1} {\lambda_1}_i$ is at least 90\% of the sum of all its eigenvalues. The same procedure is used to find $k_2$. They then define $k=\min (k_1, k_2)$. The final IS measure $\Delta$ is then the sum of the squared differences of the $k$ largest Laplacian eigenvalues: $\Delta = \sum^k_{i=1}({\lambda_1}_i - {\lambda_2}_i)^2$. The lower $\Delta$, the \emph{more} similar are the graphs and, consequently, the more isomorphic are the two embedding spaces.

\vspace{1.8mm}
\noindent\textbf{Gromov-Hausdorff Distance (GH)} It measures the worst case distance between two metric spaces $\mathcal{X}$ and $\mathcal{Y}$ with a distance function $m$ as: 

\vspace{-1mm}
{\footnotesize
\begin{align}
\begin{split}
    \mathcal{H}(\mathcal{X}, \mathcal{Y}) = \max\{& \sup_{x \in \mathcal{X}} \inf_{y \in \mathcal{Y}} m(x, y), \sup_{y \in \mathcal{Y}} \inf_{x \in \mathcal{X}} m(x, y)\}
\end{split}
\end{align}}%
\noindent It computes the distance between the nearest neighbors that are farthest apart. The Gromov-Hausdorff distance then minimizes this distance over all isometric transforms $\mathcal{X}$ and $\mathcal{Y}$:

\vspace{-1mm}
{\footnotesize
\begin{align}
    \mathcal{GH}(\mathcal{X}, \mathcal{Y}) = \inf_{f,g} \mathcal{H}(f(\mathcal{X}), g(\mathcal{Y}))
\end{align}}%
\noindent Computing $\mathcal{GH}$ directly is computationally intractable in practice, but it can be tractably approximated by computing the Bottleneck distance between the metric spaces \cite{chazal2009gromov}.

\section{Source and Target Selection Analysis}
\label{s:appendix_b}
In addition to the correlation analyses in the main text that aggregate all language pairs, some tasks and datasets also support analyses where one language is fixed as a target language (i.e., \textit{source-language selection analysis}), or as a source language (i.e., \textit{target-language selection analysis}). Such analyses could inform us on how to choose the right transfer language. That is, given a target language one would like to transfer to, which is the best source language to transfer from, and vice versa. These analyses are conducted for the following tasks with sufficient language pairs: BLI with PanLex, parsing, and POS tagging.

For these analyses we report average correlation: across target languages in the source-language selection analysis, or across the source languages in the target-language selection analysis. We provide the percentage of times each compared measure scored the highest for a particular task and setting.

Stepwise regression analysis is not suitable for the selection analysis due to the limited number of language pairs in each language selection setup (e.g., PanLex offers 14 language pairs for each source- or target- language selection analysis). These conditions impede the statistical significance power of the tests which stepwise regression requires. We therefore opt for a standard multiple linear regression model instead; the regressors include the isomorphism measure with the highest individual correlation combined with the linguistic measures. Similarly to the stepwise analysis, we report the unified correlation coefficient, $\hat{r}$. 

We observe that the same findings reported for the aggregated correlation analyses (Tables \ref{tab:full_BLI_table} and \ref{tab:downstream_table} in the main text) also hold for the language selection analyses (Table~\ref{tab:within_table} below). This indicates that our spectral measures have an applicative value as they can facilitate better cross-language transfer.

We also observe interesting patterns in the selection analyses for the POS tagging task in Table~\ref{tab:within_table}: While the results in the target-language selection analysis largely follow the main-text results, the same does not hold for source-language selection (Table \ref{tab:within_table}, POS Target and Source columns). We speculate that this is in fact an artefact of the experimental design of \newcite{lin-etal-2019-choosing}. Their set of target languages deliberately comprises only truly low-resource languages, and such languages are expected to have lower-quality embedding spaces. Transferring to such languages is bound to fail with most source languages regardless of the actual source-target language similarity. The difficulty of this setting is reflected in the actual scores: average accuracy scores for the best source-target combination is 0.55 in the source-language selection analysis, and 0.92 for target-language selection.

\section{Single and Combined Analysis}
\label{s:appendix_c}
\begin{figure}[h!]
  \centering
  \begin{subfigure}[b]{0.49\linewidth}
    \includegraphics[width=\linewidth]{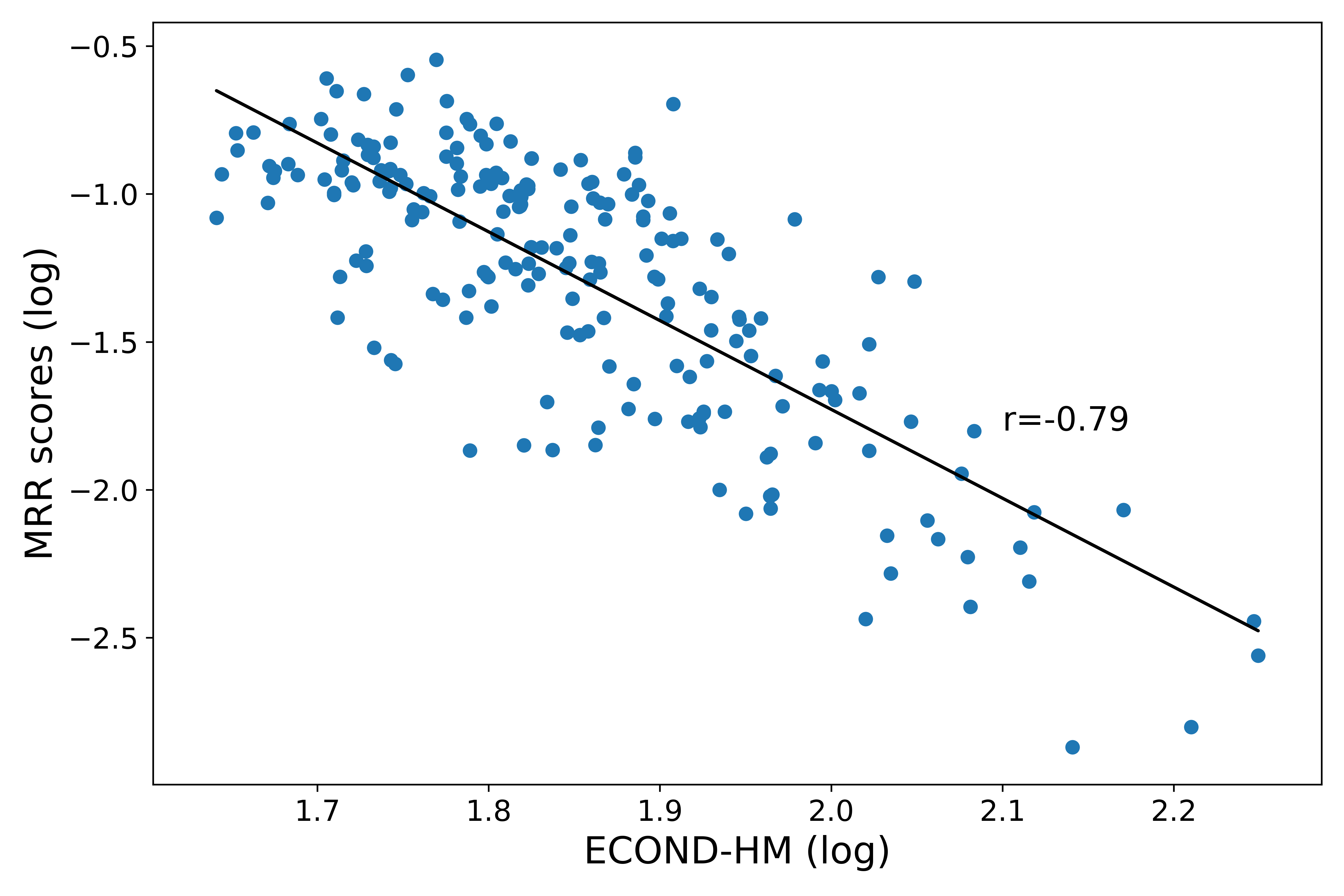}
  \end{subfigure}
  \begin{subfigure}[b]{0.49\linewidth}
    \includegraphics[width=\linewidth]{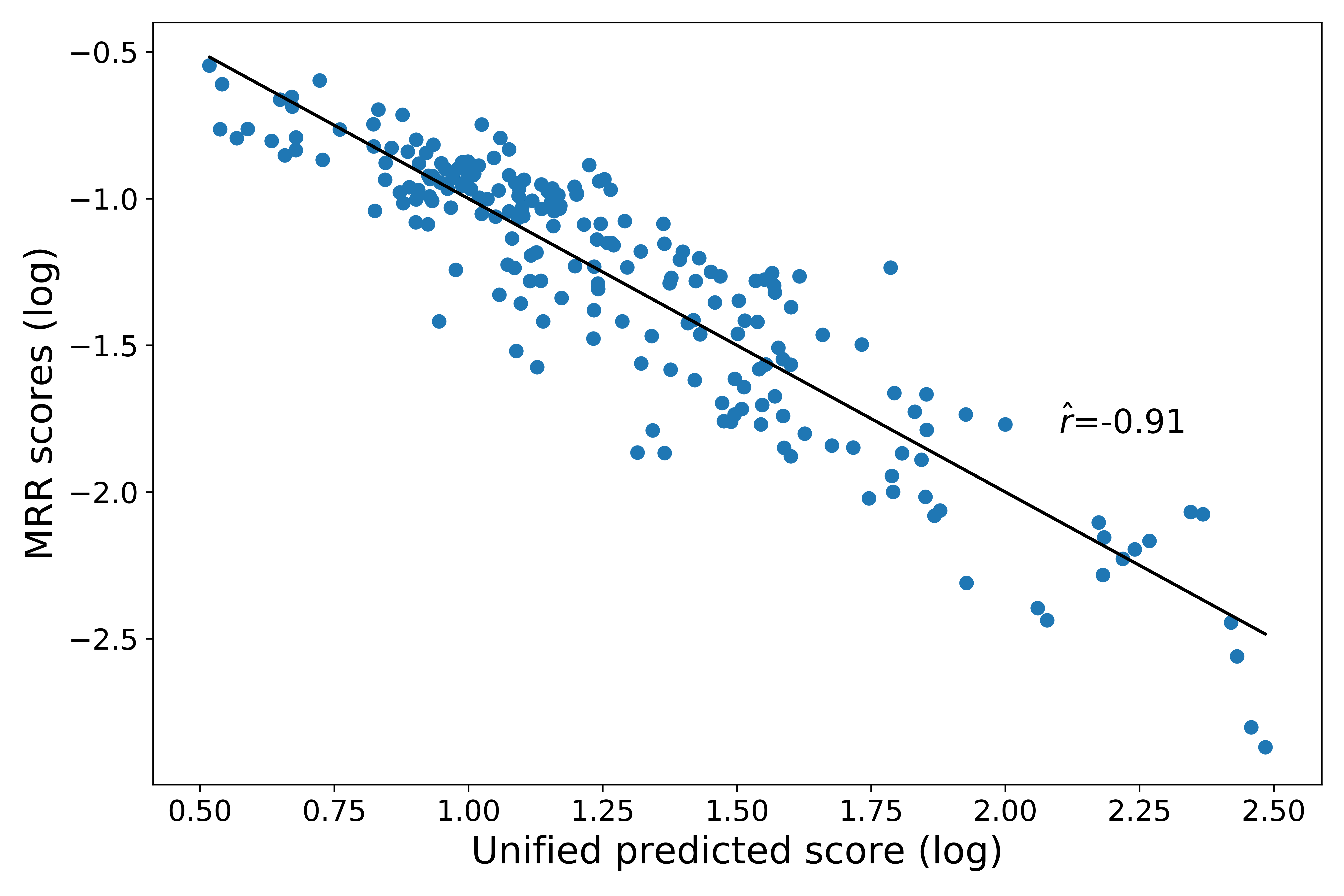}
  \end{subfigure}
  \caption{Scatter plots with least square regression lines for the \textsc{Sup} method in the PanLex BLI model (leftmost column in Table \ref{tab:full_BLI_table} of the main paper). The left panel presents results for the best single isomorphism measure, \textsc{econd-hm}. The right panel presents results for the combined unified model based on the regression analysis $\hat{r}$ that includes linguistic measures. $\hat{r}$'s sign was flipped (right panel) to make the graphs directly comparable.}
  \label{fig:single_correlation}
\end{figure}

We show (Figure \ref{fig:single_correlation}) a single experimental condition (\textsc{Sup} method in the PanLex BLI dataset, leftmost column in Table 1 in the main text) to illustrate the data distribution and the correlation for spectral-based measures (e.g., \textsc{econd-hm}), and their improvement once this measure is combined with linguistic measures through regression analysis.

\onecolumn
\begin{table}[ht!]
\centering
\def\arraystretch{1.0}
{\footnotesize
\begin{tabularx}{\linewidth}{ll lllllllll}
\toprule 
{\textit{Task}} & {} & {} & {\bf BLI} & {} & {}& {} & {\bf DEP} & {} & {\bf POS} \\ 
{\textit{Selection}} & {} & {Target} & {} & {} & {Source} & {} & {Target} & {Source} & {Target} & {Source} \\ 
{} & \textsc{Sup} & \textsc{Sup+} & \textsc{Unsup} & \textsc{Sup} & \textsc{Sup+} & \textsc{Unsup} & {} & {} & {} & {} \\ 
\midrule
{\textsc{svg}} & {-.57$^{31\%}$} & {-.56$^{31\%}$} & {\bf -.54$^{35\%}$} & {-.53$^{28\%}$} & {\bf -.54$^{29\%}$} & {\bf -.56$^{38\%}$} & {\bf -.62$^{56\%}$} & {\bf -.73$^{68\%}$} & {-.51$^{4\%}$} & {-.10} \\
{\textsc{cond-m}} & {-.59} & {-.55} & {-.42} & {-.51} & {-.49} & {-.38} & {-.44} & {-.64} & {-.58} & {-.08} \\
{\textsc{econd-m}} & {\bf -.64$^{31\%}$} & {\bf -.57$^{25\%}$} & {-.44$^{17\%}$} & {\bf -.54$^{25\%}$} & {-.53$^{24\%}$} & {-.43$^{17\%}$} & {-.51$^{3\%}$} & {-.70$^{6\%}$} & {\bf -.61$^{10\%}$} & {.05} \\

\cmidrule(lr){1-11}

{\textsc{gh}} & {-.32} & {-.33$^{3\%}$} & {-.27$^{7\%}$} & {-.31} & {-.28} & {-.26} & {-.51$^{16\%}$} & {-.63$^{13\%}$} & {-.58$^{34\%}$} & {-.11} \\
{\textsc{is}} & {-.30} & {-.30} & {-.30$^{3\%}$} & {-.33$^{3\%}$} & {-.34} & {-.32$^{7\%}$} & {-.52$^{16\%}$} & {-.58$^{10\%}$} & {-.43$^{50\%}$} & {-.23} \\
\cmidrule(lr){1-11}

{\textsc{phy}} & {-.25$^{7\%}$} & {-.27$^{3\%}$} & {-.23$^{3\%}$} & {-.41$^{24\%}$} & {-.42$^{34\%}$} & {-.32$^{10\%}$} & {-.44} & {-.42} & {-.14} & {-.14} \\
{\textsc{typ}} & {-.50$^{24\%}$} & {-.49$^{28\%}$} & {-.42$^{28\%}$} & {-.45$^{17\%}$} & {-.41$^{10\%}$} & {-.36$^{21\%}$} & {-.40$^{3\%}$} & {-.42} & {-.18$^{2\%}$} & {-.05} \\
{\textsc{geo}} & {-.29$^{7\%}$} & {-.34$^{10\%}$} & {-.32$^{7\%}$} & {-.31$^{3\%}$} & {-.32$^{3\%}$} & {-.32$^{7\%}$} & {-.48$^{6\%}$} & {-.56$^{3\%}$} & {-.12} & {\bf -.33} \\
\cmidrule(lr){1-11} 

{$\hat{r}$} & {.86} & {.83} & {.76} & {.83} & {.83} & {.77} & {.79} & {.85} & {.68} & {--} \\

\bottomrule
\end{tabularx}}%
\vspace{-0mm}
\caption{Correlation scores in \textit{source-language} (Source) and \textit{target-language} (Target) selection analyses. The best distance measure per column is provided in \textbf{bold}. The percentage of cases a measure topped the others is shown in superscript (see details in Appendix B). $\hat{r}$ refers to the unified correlation coefficient from the multiple regression model (see details in Appendix B).}

\label{tab:within_table}
\vspace{-0mm}
\end{table}

\onecolumn
\begin{table}[!t]
\centering
\def\arraystretch{0.94}
{\footnotesize
\begin{tabularx}{\linewidth}{llXXXXXX}
\toprule 
\textbf{Language family} & \textbf{Language} & \textbf{PanLex} & \textbf{MUSE} & \textbf{GTrans} & \textbf{MT} & \textbf{DEP} & \textbf{POS} \\ 
\midrule
Germanic (IE)     & English               & 28     & 88   & 7      & 54 & 62  & 17  \\
Germanic (IE)     & German                & 28     & 10   & 7      & 1  & 62  & 17  \\
Germanic (IE)     & Dutch                 & 28     & 2    &        & 1  & 62  & 17  \\
Germanic (IE)     & Swedish               & 28     & 2    &        & 1  & 62  & 17  \\
Germanic (IE)     & Danish                & 28     & 2    &        & 1  & 62  & 17  \\
Germanic (IE)     & Norwegian             & 28     & 2    &        & 1  & 62  & 17  \\
Germanic (IE)     & Afrikaans             &        & 2    &        &    &     & 65  \\
Germanic (IE)     & Faroese               &        &      &        &    &     & 50  \\
Romance (IE)      & Italian               & 28     & 10   & 7      & 1  & 62  & 17  \\
Romance (IE)      & Portuguese            & 28     & 10   &        & 1  & 62  & 17  \\
Romance (IE)      & Spanish               & 28     & 10   &        & 1  & 62  & 17  \\
Romance (IE)      & French                & 28     & 10   & 7      & 1  & 62  & 17  \\
Romance (IE)      & Romanian              & 28     & 2    &        & 1  & 62  & 17  \\
Romance (IE)      & Catalan               & 28     & 2    &        &    & 62  & 17  \\
Romance (IE)      & Galician              &        &      &        & 1  &     & 17  \\
Romance (IE)      & Latin                 &        &      &        &    & 62  & 17  \\
Slavic (IE)       & Croatian              & 28     & 2    & 7      & 1  & 62  & 17  \\
Slavic (IE)       & Polish                & 28     & 2    &        & 1  & 62  & 17  \\
Slavic (IE)       & Russian               & 28     & 2    & 7      & 1  & 62  & 17  \\
Slavic (IE)       & Czech                 & 28     & 2    &        & 1  & 62  & 17  \\
Slavic (IE)       & Bulgarian             & 56     & 2    &        & 1  & 62  & 17  \\
Slavic (IE)       & Bosnian               &        & 2    &        & 1  &     &     \\
Slavic (IE)       & Macedonian            &        & 2    &        & 1  &     &     \\
Slavic (IE)       & Slovak                &        & 2    &        & 1  & 62  & 17  \\
Slavic (IE)       & Slovenian             &        & 2    &        & 1  & 62  & 17  \\
Slavic (IE)       & Ukrainian             &        & 2    &        & 1  & 62  & 17  \\
Slavic (IE)       & Belarusian            &        &      &        & 1  &     & 65  \\
Slavic (IE)       & Serbian               &        &      &        & 1  &     & 17  \\
Hellenic (IE)     & Modern Greek          &        & 2    &        & 1  &     & 17  \\
Balto-Slavic (IE) & Lithuanian            & 28     & 2    &        & 1  &     & 65  \\
Balto-Slavic (IE) & Latvian               &        & 2    &        &    & 62  & 17  \\
Indo-Iranian (IE) & Bengali               &        & 2    &        & 1  &     &     \\
Indo-Iranian (IE) & Persian               &        & 2    &        & 1  &     & 17  \\
Indo-Iranian (IE) & Hindi                 &        & 2    &        & 1  & 62  & 17  \\
Indo-Iranian (IE) & Kurdish               &        &      &        & 1  &     &     \\
Indo-Iranian (IE) & Marathi               &        &      &        & 1  &     & 65  \\
Indo-Iranian (IE) & Urdu                  &        &      &        & 1  &     & 17  \\
Indo-Iranian (IE) & Sanskrit              &        &      &        &    &     & 50  \\
Celtic (IE)       & Irish                 &        &      &        &    &     & 65  \\
Celtic (IE)       & Breton                &        &      &        &    &     & 50  \\
Albanian (IE)     & Albanian              &        & 2    &        & 1  &     &     \\
Armenic (IE)      & Armenian              &        &      &        & 1  &     & 65  \\
Turkic            & Turkish               & 28     & 2    & 7      & 1  &     & 17  \\
Turkic            & Azerbaijani           &        &      &        & 1  &     &     \\
Turkic            & Kazakh                &        &      &        & 1  &     & 65  \\
Turkic            & Uighur                &        &      &        &    &     & 17  \\
Semitic           & Hebrew                & 28     & 2    &        & 1  & 62  & 17  \\
Semitic           & Arabic                &        & 2    &        & 1  & 62  & 17  \\
Semitic           & Amharic               &        &      &        &    &     & 50  \\
Uralic            & Hungarian             & 28     & 2    &        & 1  &     & 65  \\
Uralic            & Finnish               & 28     & 2    & 7      & 1  & 62  & 17  \\
Uralic            & Estonian              & 28     & 2    &        & 1  & 62  & 17  \\
Austronesian      & Indonesian            & 28     & 2    &        & 1  & 62  & 17  \\
Austronesian      & Malay                 &        & 2    &        & 1  &     &     \\
Austronesian      & Tagalog               &        & 2    &        &    &     & 50  \\
Dravidian         & Tamil                 &        & 2    &        & 1  &     & 65  \\
Dravidian         & Telugu                &        &      &        &    &     & 65  \\
Sino-Tibetan      & Chinese               &        & 2    &        & 1  & 62  & 17  \\
Sino-Tibetan      & Burmese               &        &      &        & 1  &     &     \\
Japonic           & Japanese              &        & 2    &        & 1  & 62  & 17  \\
Kartvelian        & Georgian              & 28     &      &        & 1  &     &     \\
Koreanic          & Korean                & 28     & 2    &        & 1  & 62  & 17  \\
Mongolic          & Mongolian             &        &      &        & 1  &     &     \\
Niger-Congo       & Yoruba                &        &      &        &    &     & 50  \\
Austroasiatic     & Vietnamese            &        & 2    &        & 1  &     & 17  \\
Tai-Kadai         & Thai                  & 28     & 2    &        & 1  &     & 50  \\
Isolate           & Basque                & 28     &      &        & 1  &     & 17  \\
Not defined       & Esperanto             & 28     &      &        & 1  &     &     \\

\bottomrule
\end{tabularx}}%
\vspace{-0mm}
\caption{Summary of all the languages included in our analyses. The numbers in each cell indicate the number of different language pairs where each language was included, per each task and dataset. IE refers to the Indo-European language group.}
\label{tab:language_counts}
\vspace{-1.5mm}
\end{table}
\twocolumn

\end{document}